\documentclass{article}

\usepackage{arxiv}
\usepackage{graphicx}
\usepackage{float} 
\usepackage{subfigure}
\usepackage[utf8]{inputenc} 
\usepackage[T1]{fontenc}    
\usepackage{hyperref}       
\usepackage{url}            
\usepackage{booktabs}       
\usepackage{amsfonts}       
\usepackage{nicefrac}       
\usepackage{microtype}      
\usepackage{cite}
\usepackage{amsmath}
\DeclareMathOperator*{\argmax}{argmax}
\usepackage{indentfirst}
\title{Mastering the working sequence in human-robot collaborative assembly based on reinforcement learning}

\author{
  Tian Yu \\
  Department of Mechanical and Aerospace Engineering\\
  University of Virginia\\
  Charlottesville, VA 22903 \\
  \texttt{ty2yy@virginia.edu} \\
   \And
  Jing Huang \\
  Department of Mechanical and Aerospace Engineering\\
  University of Virginia\\
  Charlottesville, VA 22903 \\
  \texttt{jh3ex@virginia.edu} \\
   \And
 Qing Chang\thanks{is the corresponding author} \\
 Department of Mechanical and Aerospace Engineering\\
  University of Virginia\\
  Charlottesville, VA 22903 \\
  \texttt{qc9nq@virginia.edu} \\
}

\begin{document}
\maketitle

\begin{abstract}
A long-standing goal of the Human-Robot Collaboration (HRC) in manufacturing systems is to increase the collaborative working efficiency. In line with the trend of Industry 4.0 to build up the smart manufacturing system, the Co-robot in the HRC system deserves better designing to be more self-organized and to find the superhuman proficiency by self-learning. Inspired by the impressive machine learning algorithms developed by Google Deep Mind like Alphago Zero, in this paper, the human-robot collaborative assembly working process is formatted into a chessboard and the selection of moves in the chessboard is used to analogize the decision making by both human and robot in the HRC assembly working process. To obtain the optimal policy of working sequence to maximize the working efficiency, the robot is trained with a self-play algorithm based on reinforcement learning, without guidance or domain knowledge beyond game rules. A neural network is also trained to predict the distribution of the priority of move selections and whether a working sequence is the one resulting in the maximum of the HRC efficiency. An adjustable desk assembly is used to demonstrate the proposed HRC assembly algorithm and its efficiency.
\end{abstract}

\section{Introduction}
Industrial robots have been successfully used to perform repetitive tasks with a high precision. However, there are tasks, such as complicated assembly works, that are less structured and too complex to be fully automated and thus cannot be totally performed by robots. Moreover, evaluation of the performance and the flexible adjustment by the human are sometimes necessary, which makes it impossible to fully replace humans with robots. Therefore, human–robot collaboration (HRC) systems are developed in industry to take advantages of the abilities of both humans and robots \cite{bauer2008human}. Unlike ordinary industrial robotics where the environment is structured and known, in HRC systems, the robots interact with humans who may potentially have very different skills and capabilities \cite{fong2003survey}. Over the past two decades, a significant number of researches has been done on the design of the HRC in manufacturing systems to improve the safety, quality, and efficiency of the system \cite{kruger2009cooperation,michalos2014robo,papakostas2011industrial,tan2009human,pedrocchi2013safe}. Since its declaration in late 1990s \cite{colgate1996cobots}, the collaborative robots or cobots have been playing an increasingly significant role in the HRC in manufacturing systems as the assistants for humans \cite{helms2002rob,schraft2005powermate}. However, the traditional viewpoint has been mostly focused on the development of the hardware \cite{kosuge1998human,mizoguchi1999human}, which results in machine-driven collaboration with less consideration of the function flexibility of the cobots. These studies are human-centralized, in which the cobots are limited to the scheme designed by humans and programmed based on the human domain knowledge, guidance or expert experience. Cobots have no self-learning capability, let alone the ability of self-organizing and the superhuman decision making that may surpass the existing models. Therefore, to explore a better method to accomplish a given task with the minimum workload, it is desired to develop human–robot systems in which the cobots can better cooperate with humans in a more autonomous way. The cobots will not be supposed to be regarded as only working tools anymore, but to have their own abilities to judge the system states with corresponding decision-making capability and are able to adapt themselves to various levels of the human operators.\par The assembly work in manufacturing is significantly important, which integrates parts and components to realize the final products. The application of HRC in complex, continually changing and variety-oriented assembly processes is still limited. One of the main challenges is that the HRC assembly environment is complex and dynamic. Although there might be a list of assembly tasks, the tasks are not necessarily in sequential order and many tasks can be performed in parallel, and some shared tasks can be taken by either human operators or robots. With the uncertainties involving human operator’s performance level and other random disruptions in plant floor (e.g., machine tool random failure), the decision on which tasks to be taken by what available resource, i.e. robots or human operators, and in what sequences, will have profound influence on the productivity of the assembly process. The current practice is mostly a manual process that heavily depends on human experience. The cobot is pre-programmed to perform repetitive tasks in limited assembly process. When assembly tasks change, the whole system must be reprogrammed by robotics experts, while operators working on the floor usually do not possess the expertise to reprogram the system. This requires the design of more intelligent cobot for the HRC in manufacturing assembly process.\par A large amount of the latest existing HRC studies on efficiency and safety of the system are trajectory based, focusing on the human plan recognition and the prediction of human motions\cite{modares2015optimized,cherubini2016collaborative,liu2018serocs}. Very few works have been done on finding a better decision-making policy or a better task scheduling method to improve the working efficiency of the whole system. Wilcox et al. \cite{wilcox2013optimization} provide an Adaptive Preferences Algorithm to find the optimal scheduling policy for the robot assistant to adapt to the change of human preference in the task sequence. However, their method is only demonstrated within a very small task space and with simple task structure which can be solved with mathematical optimization. For larger task spaces and more complex task structures, the application of mathematical optimization to the job dispatching problem will be Non-deterministic Polynomial-time (NP) hard.\par With the emerging of smart manufacturing of Industry 4.0 in recent years, plenty of advanced approaches have been developed in system control and decision making \cite{kang2016smart}, in which the application of machine learning (ML) is the most impressive part with increasingly powerful algorithms. For example, in scheduling, tons of research have been done using reinforcement learning (RL). In dynamic job shop scheduling (DJSS) problem, Shahrabi et al. \cite{shahrabi2017reinforcement} use RL with a Q-factor algorithm to enhance the performance of the scheduling method, considering random job arrivals and machine breakdowns. In maintenance scheduling problem, many researchers apply RL to find out the optimal maintenance policies in various manufacturing systems and obtain better system performance than traditional maintenance policies \cite{ramirez2010optimization,wang2016multi,huang2019machine}. And in resource allocation, Ou et al. \cite{ou2018gantry,ou2019simulation} have contributed a lot in the real-time scheduling of the gantry in a work cell. Although these researches are not on HRC manufacturing systems, the systematic way to formulate the decision making problem and using RL to solve the problem have laid a solid foundation in studying the human-machine interaction (such as in maintenance scheduling) and robot-machine interaction (such as in gantry scheduling). \par In the past, the traditional RL methods are commonly limited to low-dimension problems. The recent RL algorithms are becoming increasingly powerful in solving extremely high-dimension problems thanks to the advancement in deep learning. One of the most exciting researches on RL in dealing with extremely large state space decision making problem in the past few years is probably Google Deep Mind’s Alpha Go and Alphago Zero \cite{maddisonmastering,silver2017article}. In Alphago Zero, an algorithm that is based on Mote Carlo Tree Search with a deep neural network trained to improve the strength of the tree search is introduced, which results in a superhuman performance in mastering the complex game of Go. Inspired by self-play algorithm of Alphago Zero \cite{silver2017article}, we formulate the manufacturing assembly process as a chessboard game with the specific assembly rules determined by the required constraints. By integrating RL and deep neural network, we can take up the challenge to study the decision making and workload scheduling for both humans and cobots in HRC manufacturing systems. The aim of this paper is to train the robot with a self-play algorithm based on RL without guidance or domain knowledge to find an optimal policy that can maximize the assembly work efficiency and can quickly adapt to changing environment and task changes.\par The contributions of this work are in: (1) A typical assembly process is formulated as an assembly chessboard game with specific rules associated with assembly constraints. This will simplify the optimization problem, where the constraints are not needed to be considered explicitly. It also allows us to leverage the powerful RL and deep neural network algorithms to deal with extremely complicated assembly works. (2) We develop a multi-agent self-play RL algorithm to perform task scheduling, starting from random play, without any supervision or use of human data. The algorithm is inspired by Alphago Zero, with the goal changes from competition to multi-agent collaboration. The algorithm also addresses specific challenges in manufacturing assembly that do not exist in competitive game. \par The remainder of the paper is organized as following: Section 2 states the problem assumptions and objectives. In section 3, the HRC assembly problem is formulated as a RL problem by defining the states, actions and reward function. Section 4 introduces the MCTS-CNN algorithm used to obtain the optimal HRC policy. A demonstration case is presented in Section 5. Finally, conclusions and future research directions are provided in Section 6.

\section{Problem description}
\subsection{HRC assembly problem description}
The assembly job of a complex product can usually be decomposed into some number of sequential subtasks. There are two crucial matters, i.e. task allocation and task sequencing, that are directly related to the efficiency of the HRC assembly system. In order to provide some concreteness to the problem, we use the lift desk assembly job shown in Figure 1 to demonstrate these two aspects.\par
\begin{figure}[h]
  \centering
  \includegraphics[width=0.7\textwidth]{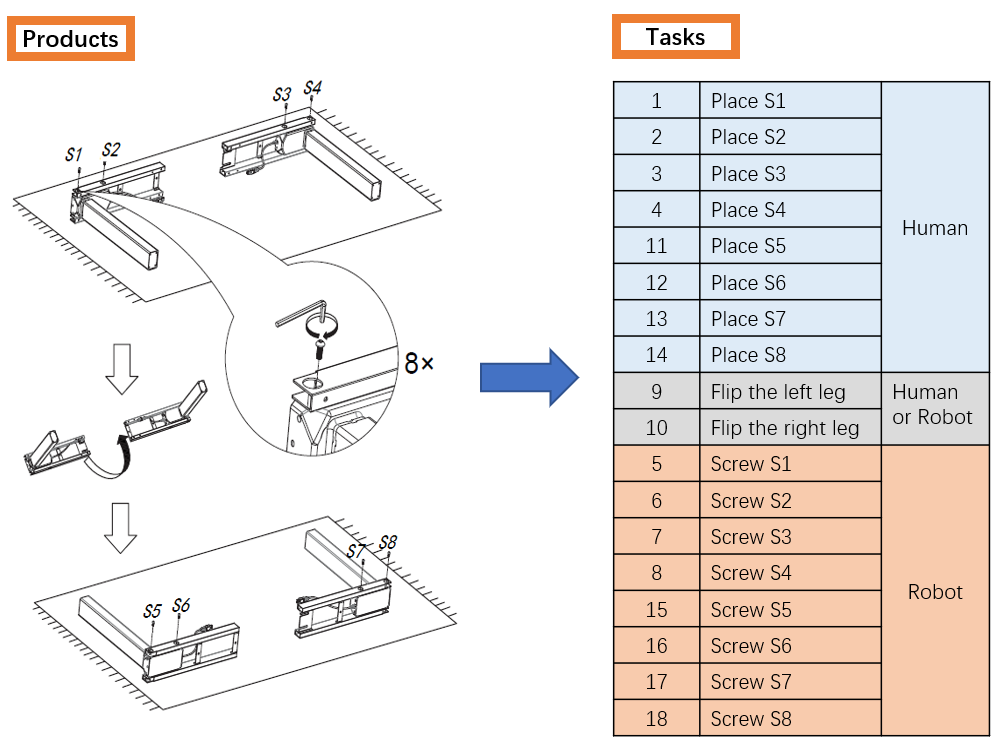}
  \caption{Diagram of decomposing the product in the human-robot collaborative assembly into tasks for humans, robots or both with a rough assembly sequence.}
  \label{fig:fig1}
\end{figure}
On one hand, either human or robot has its own distinct strengths and skills that could be exploited in the assembly job. To this end, we categorize all the tasks into three types based on their physical properties and operation characteristics, such as geometrical stability, dexterity requirement, and tolerance limits etc.\cite{bilberg2019digital}, where type I represents the tasks that can be done by humans only, type II by robots only, and type III by either humans or robots. For example, in Fig. 1, to assemble the legs, humans can place the screws in the assembly holes quickly and precisely. However, it might be hard for a robot to accomplish this task because it requires delicate adjustment to align the screws with the assembly holes. Thus, the screw placing tasks are human-only. Once the screws are placed, the robot can locate the screws and do the screwing job even faster than human. Thus, all the screwing tasks can be classified as robot-only. Other tasks, like flipping or rotating objects, can be done by either humans or robots. The task allocation refers to not only allocating human-only (or robot-only) tasks to proper human (or robot) operators, but also, more interestingly, deciding whether human or robot to take on a type III task.\par On the other hand, given a product, its assembly sequence is constrained by the product design and quality considerations etc., but these constraints are relatively loose, which means there are still many possibilities in assembly sequences for some steps. For example, in Figure 1, to assemble the legs of the desk, robot cannot screw S1 before human place it into the assembly hole. Such kind of tasks have sequential constraints, such as “Place S1 and Screw S1”, which are based on the sequence mandated by their physical properties or constrains. Other scenarios, such as which screw to be screwed first or whether the left leg or the right leg of the desk to be assembled first, have no sequence constraints, such as “Place S1, Place S2 and Place S3”. Consequently, for HRC assembly system, it is nontrivial to have an adaptive policy that determines the optimal task sequences based on real-time system states, so as to take full advantage of the flexibility of the HRC system.\par To summarize, in this paper, the problem we are solving can be described as: in a HRC assembly process, using the subtasks with existing relationships between each other as the input and choosing the task sequence and task allocation of the overall tasks as the output, seeking a method to optimize the task sequence and task allocation for both human and robots to maximize the working efficiency of the human-robot collaborative assembly.

\begin{figure}[htbp]
\centering
\subfigure[]{
\includegraphics[width=0.7\textwidth]{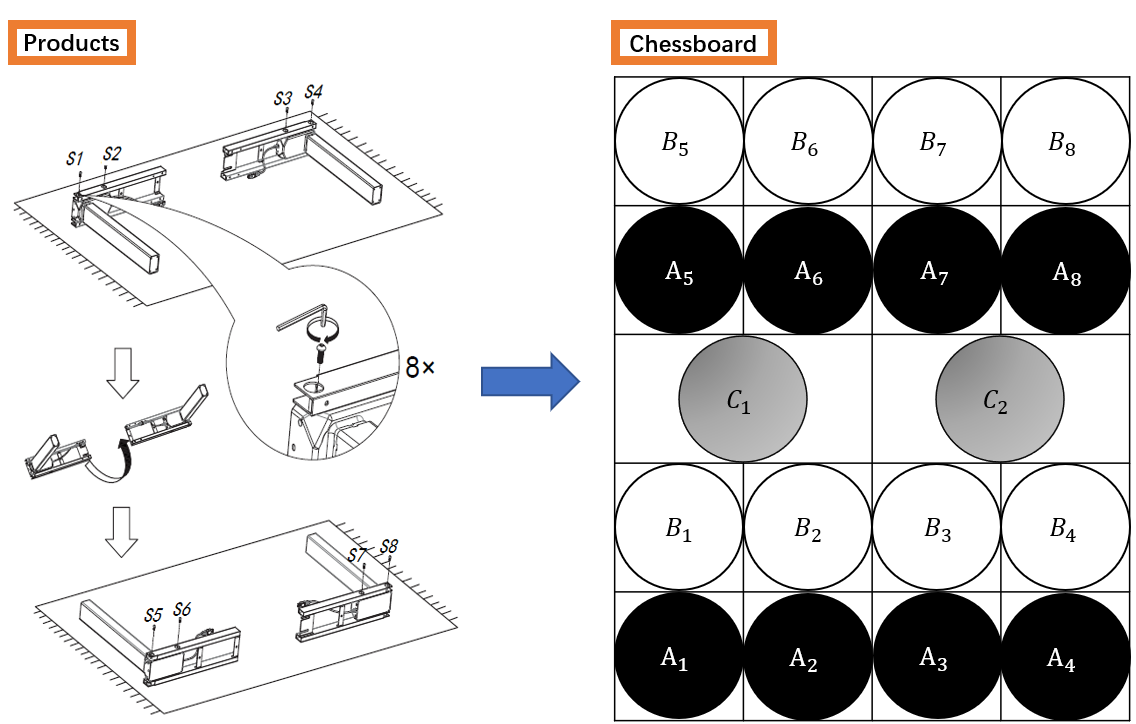}
}
\quad
\subfigure[]{
\includegraphics[width=0.7\textwidth]{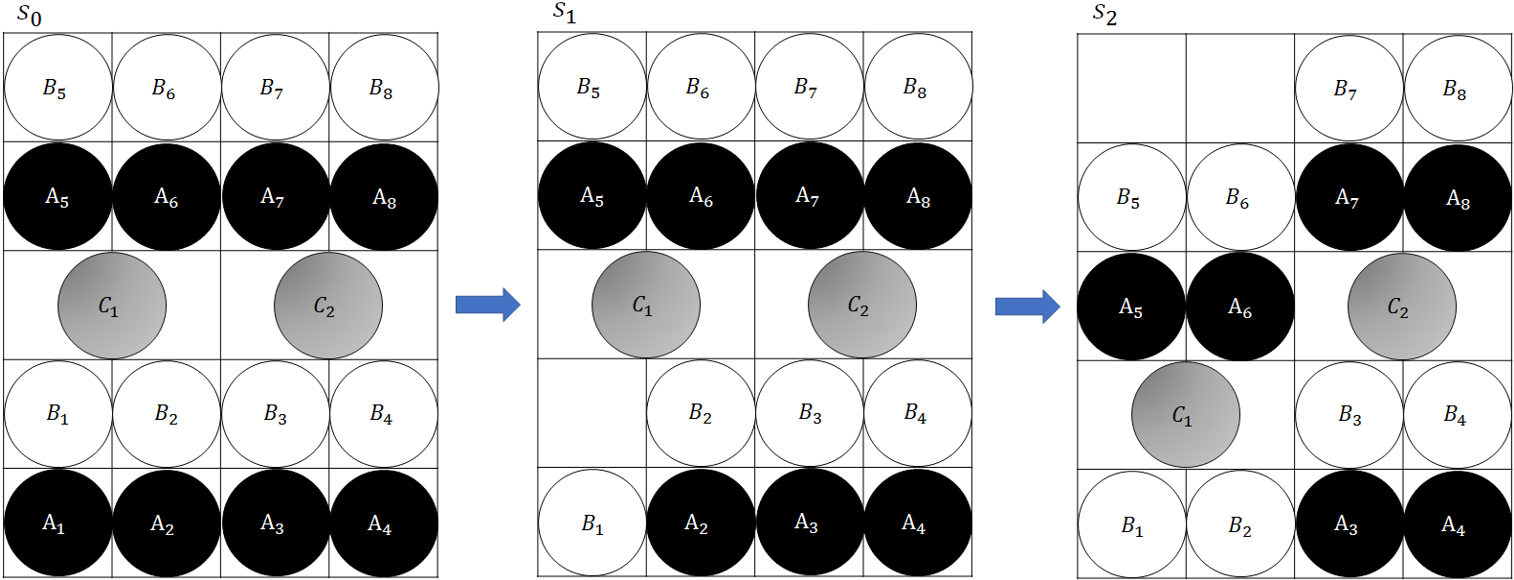}
}
\quad
\subfigure[]{
\includegraphics[width=0.7\textwidth]{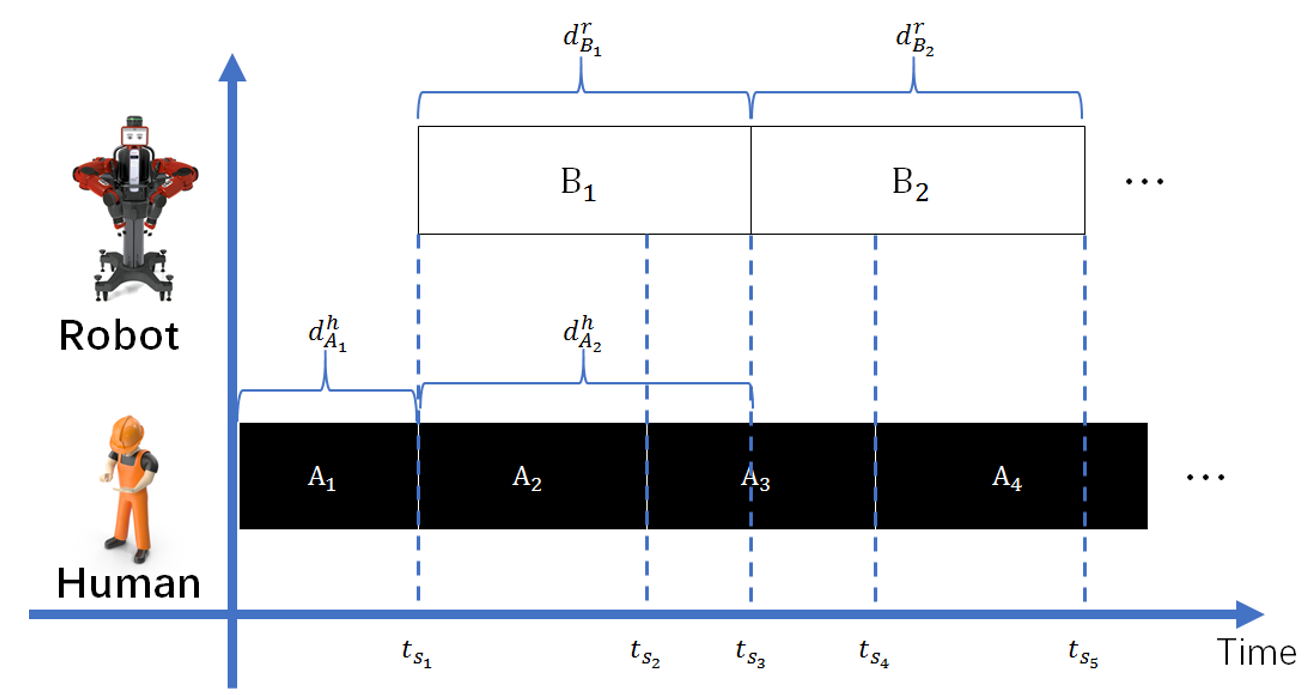}
}
\quad

\caption{Diagram of formatting the human-robot collaborative assembly process into a chessboard. (a) Diagram of mapping the assembly process in Figure 1 into a chessboard. (b) The state of the assembly process and scheme of state transitions. (c) Task sequence of both human and robot and the time of state transition.}
\label{fig:fig2}
\end{figure}

\subsection{Assembly-chessboard game}
To determine the optimal task sequence considering the constrains discussed above, we format the structure of the HRC assembly tasks into a chessboard with three different kinds of stones representing the three types of tasks as shown in Figure 2 (a). The constraints in the assembly will be represented as rules in playing the assembly-chessboard game. \par To fully reflect the task constrains and task structure complexity, the assembly-chessboard is initialized and the game rules are set as the following:
\begin{itemize}
  \item The chessboard has $w\times h$  grids, with the width $w$ determined by the number of parallel tasks and the height $h$ determined by the number of sequential tasks;
  \item Black stones denote type I tasks that can be done by humans only;
  \item White stones denote type II tasks that can be done by robots only;
  \item Grey stones denote type III tasks that can be done either by humans or robots;
  \item Tasks with sequential constraints are placed in the same column with prior tasks be set in the lower row of the chessboard;
  \item Tasks with no sequential constraints are placed in the same row of the chessboard;
  \item One stone may take more than one grid cells, depending on the number of tasks needed in that step (represented by row);
  \item Stones can only be picked from the bottom row of the chessboard;
  \item When any one stone is picked from the bottom row, the stones in all the upper rows will move down one row if the stones occupy the same size of the grid cells;
  \item The game starts with the initial pattern of the chessboard and ends until all stones are taken by humans and/or robots; 
\end{itemize}

\par Using the desk assembly as an example, the pattern of assembly chessboard is initialized shown as the first figure in Figure 2 (b). If the first task is taken by a human operator, i.e., picking a black stone $A_1$, the upper row task $B_1$ moves down to the bottom row shown as the second figure in Figure 2 (b). It means task $B_1$ can be done at the same time or at the same level of other black stone tasks and the bottom row. Note that $C_1$ cannot move down at this time, since it occupies two grid cells but only one grid cell is available. Next, if $A_2$ is picked, the $B_2$ moves down one row to fill the bottom grid cell and the available two grid cells allow $C_1$ to move down as shown in the third figure in Figure 2(b).

\section{Problem formulation}
One key advantage of formatting the assembly tasks to an assembly chessboard is that it contains not only the task structure but also the hard constrains. It offers a clear insight of the available agents and the options of tasks at any moment by just taking stones from the bottom row and following the game rules. Comparing with traditional hierarchical task tree representation which often has complicated networked structure, this representation provides a simple and accurate description of an assembly process. In this way, we can track the allocation of both humans and robots to the tasks and the sequence of their actions as shown in Figure 2(c). Therefore, our problem can be simplified as: given the pattern of the assembly chessboard at any time t when one agent or several agents are available, find the best allocation of the agents (i.e., humans or robots) to the tasks at the bottom row of the chessboard, of which the accumulated time to finish entire assembly chessboard by HRC is minimized. It is evident that this problem can be conveniently solved with the framework and algorithms in RL.

\subsection{Notations}
For the convenience of problem formulation, the following mathematical notations and assumptions are defined:
\begin{itemize}
\item $TK_u$, $u=1,2,...,U$, denotes the $U^{th}$ task in the assembly chessboard with the total number $U$;
\item $A_i$, $i=1,2,...,I$, denotes the $i^{th}$ task that can be done by humans only with the total number $I$. Each task $A_i$ has an average completion time $T_{A_i}$;
\item $B_j$, $j=1,2,...,J$, denotes the $j^{th}$ task that can be done by robots only with the total number $J$. Each task $B_j$ has an average completion time $T_{B_j}$;
\item $C_k$, $k=1,2,...,K$, denotes the $k^{th}$ task that can be done by either humans or robots with the total number $K$. Each task $C_k$ has an average completion time $T_{C_k}$;
\item $d^{h_j}_{A_i,C_k}(t)$ denotes the remaining time for a human operator $j$ to finish the tasks of $A_i$ or $C_k$. For example, if $T_{A_i}=30min$, at time $t$ it has already been done by the operator $j$ for 10 minutes, the $d^{h_j}_{A_i}(t)=20min$;
\item Similarly,$d^{h_j}_{B_i,C_k}(t)$ denotes the remaining time for a robot $i$ to finish the $B_j$ or $C_k$ task;
\item $H_i(t)\in[0,U]$, $i=1,2,...,M$ denotes the availability of the human operator $i$ at time $t$. If the human operator $i$ is working on task $TK_u$, then $H_i$ equals to $u$. Otherwise, it is set to be 0, i.e.
$ H_i(t)= \left\{ \begin{array}{ll}
0, & d^{h_i}_{TK_u}(t) = 0 \\ 
u, & d^{h_i}_{TK_u}(t) > 0 \end{array} \right. $ 
\item $R_i(t)\in[0,U]$, $i=1,2,...,N$ denotes the availability of the robot operator $i$ at time $t$. If the robot $i$ is working on task $TK_u$, then $R_i$ equals to $u$. Otherwise, it is set to be 0, i.e.
$ R_i(t)= \left\{ \begin{array}{ll}
0, & d^{r_i}_{TK_u}(t) = 0 \\ 
u, & d^{r_i}_{TK_u}(t) > 0 \end{array} \right. $ 
\item  one human and one robot are available at the same time, then the human has the priority to choose his/her action.

\end{itemize}

\subsection{Assembly-chessboard Game Problem Formulation}
The formatting of the HRC assembly process to the assembly chessboard game playing provides great convenience for the task scheduling problem to perfectly fit the Markov Decision Process (MDP) paradigm, which is the most common framework for RL. The state, action and reward function of the MDP are defined as following.\par

In our task scheduling problem, at any time $t$ the state of the chess game $s_t$ consists of two parts. One is the task information on the chessboard such as the pattern of the chessboard and the remaining time of tasks on the chessboard, and the other is the availability of the human and robot. The state $s_t$ can be defined as:
\begin{equation}
    s_t=\{p_1(t),p_2(t),...,p_Q(t),d_1(t),d_2(t),...,d_Q(t),H_1(t),H_2(t),...,H_M(t),R_1(t),R_2(t),...,R_N(t)\}
\end{equation}
where $p_i(t)=(x_i,y_i)$ is the position of the $i^{th}$ task on the chessboard; $d_i(t)\in(0,T_i]$ is the remaining time of the $i^{th}$ task; $H_i(t)\in[0,U]$ is the availability of the $i^{th}$ human operator and $R_i(t)\in[0,U]$ is the availability of the $i^{th}$ robot.\par 
For each state $s_t$, the tasks in the bottom row of the chessboard consist the options of action ${A(t),B(t),C(t)}$, in which $A(t)$ is the set of human-only tasks; $B(t)$ for robot only, and $C(t)$ for human-or-robot. Therefore, the actions of all humans and robots at time $t$ can be represented as:

\begin{equation}
    a(s_t)=\{a_t^{h_1},a_t^{h_2},..,a_t^{h_M},a_t^{r_1},a_t^{r_2},...,a_t^{r_N}\}
\end{equation}

in which, $a^{h_i}_t\in\{A(t)\cup C(t), \emptyset\}$, where $a^{h_i}_t=\emptyset$ means the human operator $i$ takes no action at time $t$, and $a^{r_i}_t\in\{B^t_j\cup C^t-a^h_t\cap C^t, \emptyset\}$ where $a^{r_i}_t=\emptyset$ means the robot $i$ takes no action at time $t$.\par

The ultimate goal of our problem is to minimize the job completion time. Bear this in mind, the reward function $r_t$ should be directly related to the elapsed time between two consecutive states. Therefore, the reward function is defined as:
\begin{equation}
    r_t=-\min_{m=1,\dots,M,n=1,\dots,N}\{d^{h_m}_{A_i,C_k}(t),d^{r_n}_{B_j,C_k}(t)\}
\end{equation}\par

From the initial state $s_0$ to the terminal state $s_{end}$, all the taken actions consist an action route $l$ denoted as $\boldsymbol{a}_l=\{a(s_0),\dots ,a(s_t),\dots ,a(s_{end})\}$ as shown in Figure 2(c). Our goal is to find the optimal HRC policy to maximize the accumulated reward $r^l=\sum_{t=0}^{end} r_t$, so as to minimize the total job completion time.\par

It is noted that for route $l$, there are $O_l=\prod^{end}_{t=0}O_{s_t}$ possible combinations of humans’ and robots’ actions, in which $O_{s_t}$ is the combinations of humans’ and robots’ actions in state $s_t$. For the example of the desk assembly, this $O_l$ can be a huge number just like in the game of chess, the options of moves are around $35^{80}$. For more complicated assembly jobs, such as automotive assembly, this can be an even larger number. Therefore, the assembly-chessboard game problem has an ultra-high dimension. A proper algorithm has to be developed to solve this problem both effectively and efficiently.

\section{Reinforcement learning for HRC assembly}

We use a specially-designed chessboard game to describe the HRC assembly process. Based on the problem formulation in previous section, it has quite a few similarities to traditional board games, e.g. the game of Go. For example, it also has finite moves, and one agent’s action inevitably affect the other’s situation. Therefore, it is possible to leverage the successful solutions to traditional board games to solve our assembly-chessboard problem. In this section, we present the MCTS-CNN algorithm used in Alphago Zero to obtain the optimal HRC policy.

\subsection{MCTS in the HRC assembly-chessboard}

In the MCTS algorithm, the information of the system state $s_t$ defined above is saved in each node $s_t$  of the searching tree. For all legal actions $a_t\in a(s_t)$, the corresponding edges $(s_t,a_t)$ are connected to the node $s_t$. The information saved in each state is a set as:
\begin{equation}
    \{N(s_t,a_t),W(s_t,a_t),P(s_t,a_t)\}
\end{equation}
in which,$N(s_t,a_t)$ is the visit count of the edge, $W(s_t,a_t)$ is the total action value and $P(s_t,a_t)$ is the probability in searching the edge $(s_t,a_t)$. The searching process always starts from the root node $s_0$ of the searching tree and ends until a leaf node $s_L$ is reached. And the action a is selected in the state $s_t$ based on a variant of the PUCT algorithm\cite{silver2017article}, which can be defined as:
\begin{equation}
    a=\argmax_{a_t}(Q(s_t,a_t)+U(s_t,a_t))
\end{equation}
in which, $Q(s_t,a_t)=W(s_t,a_t)/N(s_t,a_t)$ is the state value and $U(s_t,a_t)=cP(s_t,a_t)\sqrt{\sum_{b_t}N(s_t,b_t)}/(1+N(s_t,a_t))$ is the exploration part to balance the exploitation part which is $Q(s_t,a_t)$, $c$ is a constant that determines the level of exploration. When the leaf node is reached, it is always expanded and evaluated. Then, a reward $v$ is backed up through each previous edge of the searching route. The visit count of these edges will be incremented as $N(s_t,a_t )=N(s_t,a_t )+1$ and the action value will be updated as $W(s_t,a_t )=W(s_t,a_t )+v$. After a number of searches from the root, the agent takes an action in the root state $s_0$ based on the best policy $\pi_{opt}(a|s_0)=\max_{a_t}(N(s_0,a_t)/\sum_{b_t}N(s_0,b_t))$ proportional to the edge’s visit count. Then, after the action is done, the system will transit to a new state $s_1$. Use this new state $s_1$ as the new root node and the subtree below node $s_1$ is retained as the new searching tree. All other parts of the previous searching tree are discarded. In this way, we can find the optimal policy for the agent scheduling in each state.

\subsection{CNN in the HRC assembly chessboard}

The assembly-chessboard game has an immense state space, and therefore the deep neural network helps tackle the dimensional issue with its powerful approximation ability. In addition, when the assembly job is represented with the chessboard, then the spatial correlation within the board is of great significance. Therefore, CNN is an ideal candidate to work with MCTS to solve the assembly-chessboard problem\cite{silver2017article}.\par
\begin{figure}[h]
  \centering
  \includegraphics[width=0.7\textwidth]{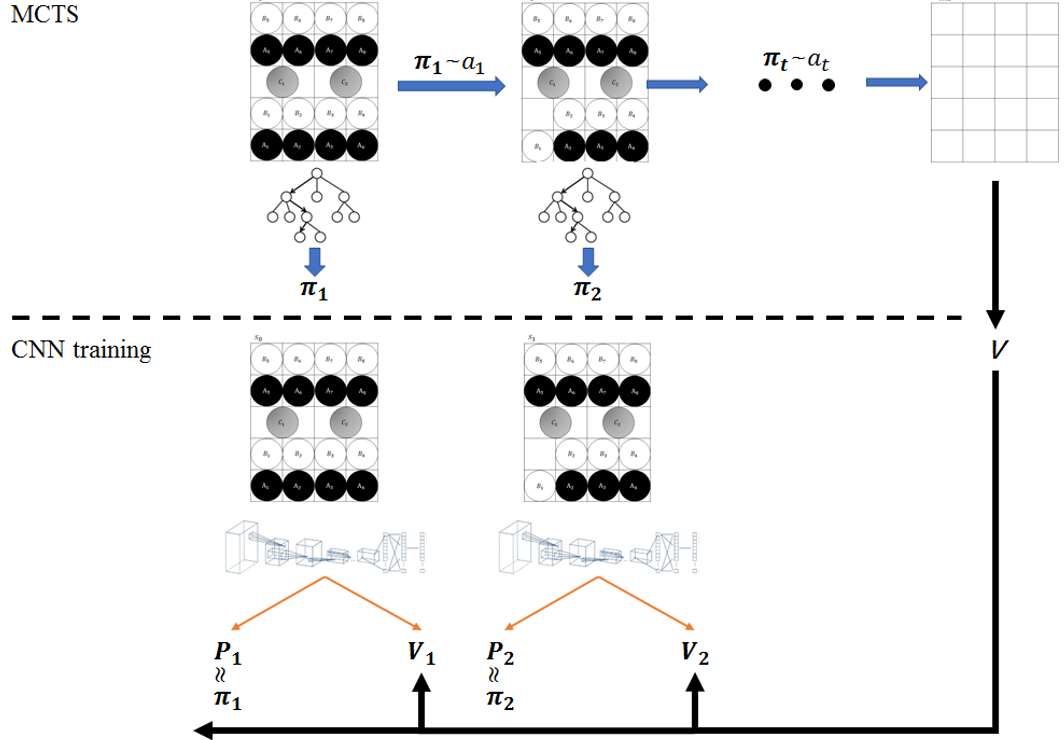}
  \caption{Diagram of the Monte Carlo Tree Search and the Convolutional Neural Network training the assembly chessboard.}
  \label{fig:fig3}
\end{figure}

The CNN $f_{\theta} (*)$ takes all the information stored in the raw chessboard representation s as input, and outputs both move probabilities and a state value, i.e. $(p,v)=f_{\theta}(s)$. The vector of move probabilities $p$ represents the probability of selecting each action. The value $v$ is a scalar evaluation, estimating the task completion time starting from current state s. This neural network combines the roles of both policy network and value network into a single architecture.\par

The CNN is trained with the data gained from MCTS. In each state $s_t$, the MCTS search outputs policies $\pi$ of taking actions. The actions selected by these policies are usually stronger than those selected from the CNN with the raw probabilities. Therefore, MCTS can be regarded as a powerful policy improvement operator. The main idea of our reinforcement learning algorithm is to use these search operators repeatedly in a policy iteration procedure, in which the neural network’s parameters are updated to make the move probabilities and the state value $(p,v)=f_{\theta}(s)$ more closely match the improved search probabilities and the real task completion time; these new parameters are used in the next iteration$(p,v)=f_{\theta}(s)$ to more closely match the improved search probabilities and the real task completion time. Figure 3 illustrates the training pipeline.

\section{Implementation and demonstration}
With the understanding of rules in the chessboard, we choose to assemble a height adjustable standing desk shown as an example. The job types are defined based on the following rational:

\begin{itemize}
    \item Matching and placing work can only be done by humans;
    \item Screwing, drilling and gumming work can only be done by robots;
    \item Flipping and rotating work can be done by either humans and robots.
\end{itemize}

\par There are 27 robot-only tasks, 19 human-only tasks, and 4 human-or-robot tasks. After fitting the tasks into the chessboard according to the rough assembly sequence, an assembly chessboard is obtained with height $h=$ and width $w=8$. There is one robot operator and one human operator cooperating with each other to complete the assembly job. The neural network architecture used in this case is defined as the followings:

\begin{itemize}
    \item Input layer with size $h\times w\times d=15\times 8\times 3$;
    \item Convolutional layer with 10 filters of kernel size $2\times 2$ with stride 1 and Relu activation;
    \item Max-pooling layer with size $2\times 2$;
    \item Convolutional layer with 10 filters of kernel size $2\times 2$ with stride 1 and Relu activation;
    \item Max-pooling layer with size $2\times 2$;
    \item Flattening layer;
    \item Dense layer of 128 units with Relu activation;
    \item First output from the dense layer: classification layer of size w with softmax activation function, and each indicates the probability choosing the $w^th$ task;
    \item Second output from the dense layer: a single regression variable predicting the state value.
\end{itemize}
 \begin{figure}[h]
  \centering
  \includegraphics[width=0.7\textwidth]{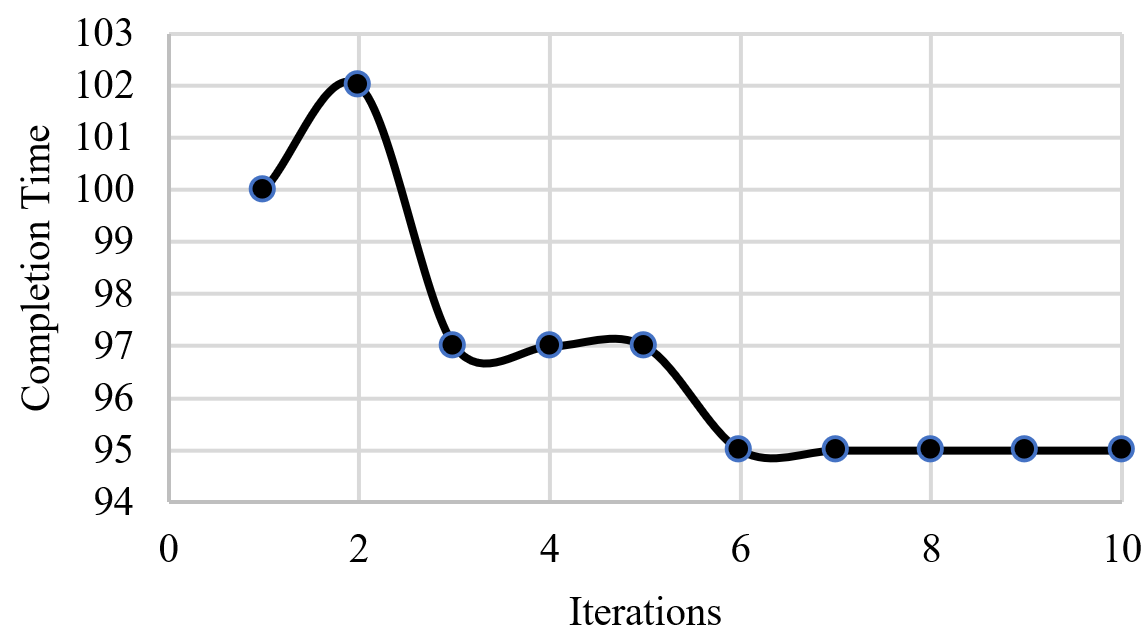}
  \caption{Diagram of the training process.}
  \label{fig:fig3}
\end{figure}   
\par For MCTS, the maximum search depth is set to be 3 and maximum searches is limited to 30 times for one root node. The parameter for the UCT algorithm is chosen to be $c=100$. We run the program in a PC with Intel Core I5-8400, UHD Graphics 630 and 12G RAM. The training progress is as shown in Figure 4. It can be seen that the proposed algorithm steadily makes progress despite of some fluctuations during the first few iterations. The shortest completion time is reached at the sixth iteration. After the sixth iteration, the completion time remains steady as 95, which indicates that the convergence has been reached.\par

For comparison purpose, two other approaches are considered in this case study. One is the exhaustive search, in which we traverse all possible routes; while the other one is to randomly choose 1,000 trajectories to finish the assembly job. It turns out that the exhaustive search is not feasible, since the trajectories possibilities explodes and finally depletes the PC memory. This situation is almost inevitable for a large-scale planning problem, since the state space explodes exponentially with decision steps. The exhaustive search program abruptly stops at step 55, and 5 more steps are yet to be traversed. We plot the total route numbers and computing time against the decision steps in Figure 5.

\begin{figure}[h]
  \centering
  \includegraphics[width=0.7\textwidth]{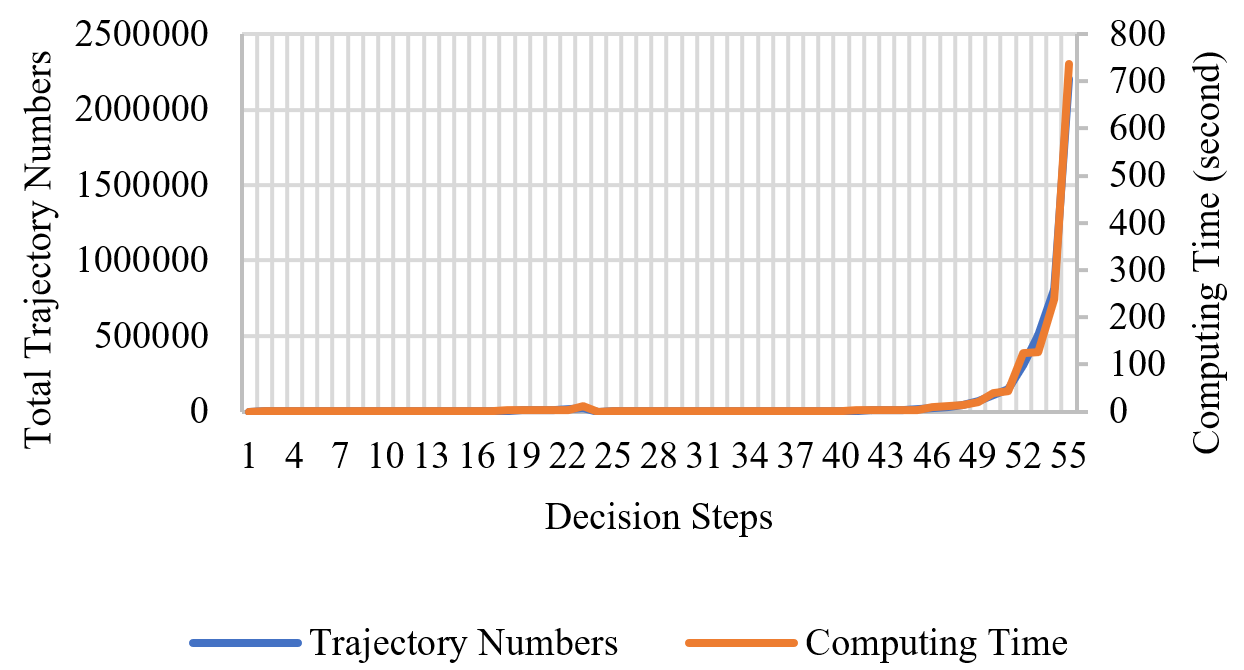}
  \caption{The computing time before depleting memory in exhaustive search.}
  \label{fig:fig3}
\end{figure}

As for the other comparison approach, we limit the possible trajectories to 1,000 by randomly sampling to make it computationally feasible. This approach mimics the scenario when both human operator and robot operator randomly choose tasks without considering the goal of minimizing the job completion time. However, as shown in Figure 6, the result reveals that the shortest completion time is hardly achieved as only one out of 1,000 gives the completion time of 95. The average job completion time is 100.58, which is far more than the shortest completion time obtained by the proposed algorithm in this paper. Therefore, this approach cannot guarantee the optimality of the randomly sampled routes. In conclusion, the proposed algorithm in this paper is effective in the online scheduling of assembly tasks in HRC scenarios.
\begin{figure}[h]
  \centering
  \includegraphics[width=0.6\textwidth]{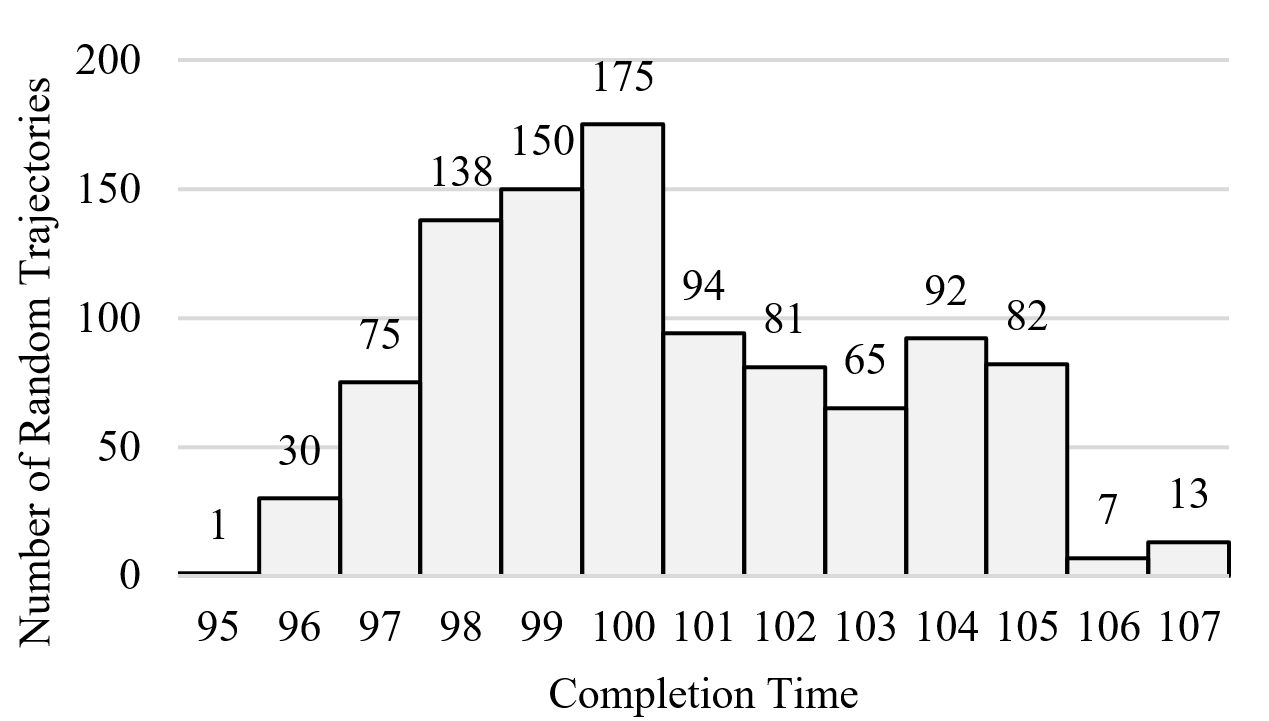}
  \caption{Histogram of the completion time with 1000 trajectories in total. }
  \label{fig:fig3}
\end{figure}

\section{Conclusion}
In this paper, we format the human robot collaborative assembly process into an assembly chessboard with the game rules. Be benefit from this formatting, we formulate the task scheduling problem in the human robot collaborative assembly process into a RL problem. The MCTS-CNN algorithm based on Alphago Zero that combines MCTS combined and CNN is used to solve the problem. A demonstration of assembling a desk with HRC has been implemented which shows the time-saving property in finding the optimal task scheduling policy and the effectiveness in improving the collaborative working efficiency of the HRC system. In our future work, more complicated scenarios such as robot random failure and the uncertainty of humans will be included into consideration and the safety of the HRC will be another key point to be investigated while improving the working efficiency of the HRC system. 

\section{Acknowledgement}
This work is supported by U.S. National Science Foundation under Grant CMMI 1351160 and 1853454.

\bibliographystyle{unsrt}  
\bibliography{references}

\end{document}